# Stable and Compact Face Recognition via Unlabeled Data Driven Sparse Representation-Based Classification

Xiaohui Yang[1], Zheng Wang[1], Huan Wu[1], Licheng Jiao[2*], Yiming Xu[3], Haolin Chen[4]

1. Henan Province Engineering Research Center for Artificial Intelligence Theory and Algorithm, School of Mathematics and Statistics, Henan University, Kaifeng, China
2. Key Laboratory of Intelligence Perception and Image Understanding of Ministry of Eduction, Xi'an, 710071, China
3. Telecom Paris, Institut Polytechnique de Paris, Paris, 91120, France
4. School of Computer and Information Engingeering, Henan University, Kaifeng, China

**Abstract**: Sparse representation-based classification (SRC) has attracted much attention by casting the recognition problem as simple linear regression problem. SRC methods, however, still is limited to enough labeled samples per category, insufficient use of unlabeled samples, and instability of representation. For tackling these problems, an unlabeled data driven inverse projection pseudo-full-space representation-based classification model is proposed with low-rank sparse constraints. The proposed model aims to mine the hidden semantic information and intrinsic structure information of all available data, which is suitable for few labeled samples and proportion imbalance between labeled samples and unlabeled samples problems in frontal face recognition. The mixed Gauss-Seidel and Jacobian ADMM algorithm is introduced to solve the model. The convergence, representation capability and stability of the model are analyzed. Experiments on three public datasets show that the proposed LR-S-PFSRC model achieves stable results, especially for proportion imbalance of samples.

**Keywords:** Sparse representation-based classification; unlabeled data; sparsity; low-rank subspace structure; face recognition

# 1 Introduction

Face recognition is an important biometric recognition method. Compared with other biometric recognition methods, face recognition has the following advantages: difficult to replicate, easy access and non-invasive. A face recognition system mainly depends on effective feature representation and classification. J. Wright et al. [1] proposed a sparse representation based classification (SRC), where the label can be assigned by class-specific reconstruction residual and sparse signal representation theory was applied to face recognition. However, SRC is limited to sufficient labeled samples per category.

Some improved SRC methods have been presented. Extended SRC (ESRC) [2] constructed an auxiliary intraclass variant dictionary which took account of differences between samples. Zhang et al. [3] revealed that collaborative representation made SRC powerful rather than $l_1$ regularization. Cai et al. [4] proposed probabilistic collaborative representation-based classification, which gave the intrinsic classification mechanism of SRC and CRC. Looking at this from another perspective, classification is to separate samples located in different linear subspaces. Liu et al. [5] proposed the low-rank representation which was a strong performer technique to analyze data drawn from multiple subspaces. Du et al. [6] incorporated low-rank and sparse constrains to represent data. Similar to SRC, however, these improved SRC methods often suffer from few effective labeled samples. Recently, deep neural networks have attracted more interest with applications in face recognition Lopes et al. [7] proposed a simple solution for facial expression recognition that used a combination of convolutional neural network and specific image pre-processing steps. Chen et al. [8] proposed softmax regression-based deep sparse autoencoder network for facial emotion recognition to address the problems of learning efficiency and computational complexity. PCANet [9], an emerging deep learning algorithm, is considered as a simple but highly competitive baseline for face recognition. However, the success of deep leaning methods is usually based on large amount of data, advanced hardware and complex structure. Generally speaking, most of the improved SRC methods and deep learning-based methods still rely on enough labeled samples. However, unlabeled data is usually much more than labeled data. Therefore, how to make efficient use of labeled and unlabeled data has become a research hotspot [10,11].

Our previous work [12] proposed a pseudo-full-space representation (PFSR) based classification (PFSRC) model for robust face recognition. PFSRC projects each labeled sample onto a pseudo-full space consisted of all the samples except itself. PFSRC aims to explore complementary information contained in available samples. Compared with standard SRC, there are some differences. Firstly, each unlabeled sample is linearly expressed by all labeled samples in standard SRC, while the PFSRC projects each labeled sample onto the pseudo-full-space via an inverse projection approach. Secondly, the matched classification criterion of standard SRC is the minimum reconstruction error, while a simple and robust decision rule, category contribution rate (CCR), is defined to match the PFSR and complete classification. Finally, PFSRC stably and effectively exploits complementary information between labeled and unlabeled samples. In PFSRC, the $l_2$ regularizer yields smooth solution, but it does not possess the sparse property. In addition, PFSRC does not mine the subspace structure of data.

For sparse representation models, different constraints can be appended according to different prior information. Sparsity characterized by $l_1$ norm [13-15] is an important prior information for recognition. Low-rankness characterized by nuclear norm is another prior information, which helps to explore the multiple subspace structures of data [16,17]. Sparse and low-rank constraints can enhance the representation and discrimination of model. Alternating direction method of multipliers (ADMM) [18] is a common used algorithm for solving two-variable convex optimization problems. The convergent solution is obtained by alternating iteration of two variables. However, the convergence of ADMM cannot be guaranteed when there are more than two variables. Fortunatly, the convergence of extension of ADMM for multi-block convex minimization problems is proved when any two coefficient matrices in objective function are orthogonal [19]. In addition, a mixed Gauss-Seidel and Jacobian ADMM (M-ADMM) [20] is proposed to complement each other, and the convergence bound is further given to show the convergence speed of the optimization algorithm. M-ADMM divides multi-block variables into two super blocks, and then updates them in sequence as Gauss-Seidel ADMMs, while the variables in every block are updated in order as Jacobian ADMMs.

Motivated by these works, a stable and compact model is proposed by imposing the low-rank and sparse constraints on the PFSR model, named LR-S-PFSR model. Without causing confusion, the corresponding LR-S-PFSR based classification is called LR-S-PFSRC. The main differences between the proposed LR-S-PFSRC and our previously published method (PFSRC) [12] are as follows.

(1) Compared with PFSRC with $l_2$ contraint, LR-S-PFSRC focuses on obtaining the sparsest and lowest-rank representation via simple linear regression and without learning, which can be achieved by imposing sparse and low rank contraints on representation matrix according to the inherent properties of data. Moreover, we exploit the discriminative nature of sparse representation to perform classification just like [1]. Meanwhile, we exploit low rank representation to recover the subspace structures from data.

(2) PFSRC is a sparse representation model with $l_2$ contraint, where the analytical solution can be obtained directly. However, the optimization of LR-S-PFSRC has the following difficulties. Firstly, it is a multivariate optimization problem, and the commonly used ADMM algorithm cannot guarantee its convergence. Secondly, since the proposed model contains the constraints $Z_{ii}=0$ and $A_{ii}=0$, projection operators need to be constructed to ensure $Z_{ii}=0$ and $A_{ii}=0$. Meanwhile, LR-S-PFSRC not only analyzes the convergence, but also shows the convergence speed through the convergence bound theorem.

(3) The proposed LR-S-PFSRC model has been verified to suitable for few labeled samples, and proportion imbalance between labeled samples and unlabeled samples.

The proposed method is an improved SRC method, which focuses on casting recognition problem as simple linear regression. Similar to SRC [1], the experiments in this paper are confined to human frontal face recognition with illumination, expressions, occlusion and disguise. The main contributions of LR-S-PFSRC are as follow.

(1) The LR-S-PFSR model further improves the performance of the PFSR by exploiting the sparsity and subspace structures of data.

(2) The M-ADMM is introduced to solve the LR-S-PFSRC model, where a pair of orthogonal matrices is verified to exist and two projector operators are constructed.

(3) The performance of the model is discussed, including representation capability and classification stability.

(4) A LR-S-PFSRC model-based robust face recognition technique is proposed to focus on problems with few labeled samples and proportion imbalance between labeled samples and unlabeled samples.

The rest of this paper is organized as follows. Section 2 describes the methodology of LR-S-PFSR model including construction and optimization. Second 3 discusses representation capability and classification stability. The robust face recognition algorithm based on LR-S-PFSRC is proposed in Section 4. Section 5 shows the experimental results. Finally, conclusions are conducted in Section 6.

## 2 The LR-S-PFSR model

After reviewing standard sparse representation and PFSR, the construction of the LR-S-PFSR model are discussed in detail. And then, the model is solved by the M-ADMM, where a pair of orthogonal matrices is verified to exist and two projector operators are constructed.

### 2.1 Standard sparse representation and PFSR

Suppose $X = [X_1, \cdots, X_j, \cdots, X_c] \in R^{d \times s_c}$ is a labeled samples set, $X_j = [x_{s_{j-1}+1}, \cdots, x_{s_j}] \in R^{d \times (s_j - s_{j-1})}$ are the $j$-th category samples, $j = 1, \cdots, c$ is the number of category, $s_j$ is the total number of labeled samples in the $j$-th category. $Y = [y_1, y_2, \cdots, y_m] \in R^{d \times m}$ is a unlabeled samples set. In SRC, each unlabeled sample $y_l \in R^d$ can be linearly represented by all labeled samples,

$$y_l = \gamma_{l,1}x_1 + \cdots + \gamma_{l,i}x_i + \cdots + \gamma_{l,s_c}x_{s_c} = \sum_{i=1}^{s_c}\gamma_{l,i}x_i = X\gamma_l, \qquad (1)$$

where $\gamma_l \in R^{s_c}$ are the corresponding coefficient vector. Without causing confusion, the corresponding projection way and representation space of SRC are called positive projection and positive space.

PFSR [12] aims to seek a representation space as large as possible, where each labeled sample $x_i$ is represented by its corresponding pseudo-full-space $V_i = \{X,Y\} - \{x_i\}$, $i = 1,\cdots,s_c$. The projection way is called inverse projection, which is opposite to that of standard sparse representation.

$$x_i = \cdots + \alpha_{i,i-1}x_{i-1} + \alpha_{i,i+1}x_{i+1} + \cdots + \beta_{i,1}y_1 + \cdots + \beta_{i,m}y_m, \qquad (2)$$

where $\alpha_{i,s} \in R$ and $\beta_{i,l} \in R$ are the corresponding coefficients before labeled samples and unlabeled samples respectively, $i = 1,2,\ldots,s_c$, $s = 1,2,\ldots,s_c$, $s \neq i$, $l = 1,2,\ldots,m$. Obviously, $V_i$ provides richer information than the labeled samples space because of the addition of unlabeled samples.

Let $Z_i = [\alpha_{i,1},\cdots,\alpha_{i,s_c},\beta_{i,1},\cdots,\beta_{i,m}]^T \in R^{(m+s_c)}$, $x_i$ can be rewritten as $x_i = V_i Z_i$. All labeled samples can be linear represented as $X = VZ$, where $V = [V_i]$ and $Z = [Z_i] \in R^{(m+s_c) \times s_c}$ are the pseudo-full-space and the corresponding coefficient matrix, respectively.

It is worth mentioning that PFSR seeks the representation space as large as possible to fully exploit the complementarity between labeled and unlabeled samples. However, for standard sparse representation, adding unlabeled samples to its representation space means that the proportion of labeled samples is reduced and the representation performance will go bad. Without causing confusion, SRC with adding unlabeled samples is called SRC+test. In Subsection 5.4.1, SRC and SRC+test are analyzed and compared. Also, we discuss the classification accuracy of SRC and PFSRC with increasing the number of unlabeled samples per category, where the premise is that the number of labeled samples per category is fixed.

## 2.2 Construction of the LR-S-PFSR model

In this subsection, we aim to seek a low-rank and sparse representation of PFSR model. The sparser the recovered representation coefficient is, the easier will it be to accurately determine the identity of the represented sample [1]. Low-rankness can reveal subspace structures information which can enhance the representation capability of model for classification tasks. Therefore, the PFSR model can be improved by introducing the sparse and low-rank constraints, and then the LR-S-PFSR model is formed. Obviously, the difference between PFSR and LR-S-PFSR is that the constraints are different, then the representation coefficients obtained are different.

As described in Subsection 2.1, the labeled samples set and the pseudo-full-space are $X$ and $V$, respectively. The optimization problem can be formulated as

$$\min_{Z} \ rank\|Z\| + \lambda_1 \|Z\|_0, \\ s.t. \ X = VZ, \ Z_{ii} = 0, \quad (3)$$

where $\lambda_1$ is is a trade-off between $rank\|Z\|$ and $\|Z\|_0$, and $Z_{ii} = 0 \ (i = 1, 2, \cdots, s_c)$ indicates that the diagonal of the coefficient matrix corresponding to the labeled samples in PFSR are 0. Assuming the labeled samples set has been arranged according to the category order, each labeled sample is linearly expressed by the pseudo-full-space which is the full space except the represented sample itself. So, the diagonal coefficients of the coefficient matrix corresponding to the labeled samples are all zeros.

$$\begin{cases} x_1 = \overbrace{0x_1 + \alpha_{1,2}x_2 + \ldots + \alpha_{1,s_c}x_{s_c}}^{\text{labeled data}} + \overbrace{\beta_{1,1}y_1 + \beta_{1,2}y_2 + \cdots \beta_{1,m}y_m}^{\text{unlabeled data}} \\ x_2 = \alpha_{2,1}x_1 + 0x_2 + \ldots + \alpha_{2,s_c}x_{s_c} + \beta_{2,1}y_1 + \beta_{2,2}y_2 + \cdots \beta_{2,m}y_m \\ \ldots \\ x_{s_c} = \alpha_{s_c,1}x_1 + \alpha_{s_c,2}x_2 + \ldots + 0x_{s_c} + \beta_{s_c,1}y_1 + \beta_{s_c,2}y_2 + \cdots \beta_{s_c,m}y_m \end{cases}$$

Fig. 1. The representation way of PFSR and LR-S-PFSR models.

Fig. 1 gives a motivating example. It is worth noting that all the PFSR-based methods are represented in this way, while the coefficients obtained are different due to different constrains.

As the optimization problem Eq. (3) is non-convex, the $l_0$ norm is substituted by $l_1$ norm and the rank is substituted by the nuclear norm [16]. Hence, the optimization problem Eq. (3) is relaxed as

$$\min_{Z} \|Z\|_* + \lambda_1 \|Z\|_1,$$
$$s.t. \ X = VZ, \ Z_{ii} = 0.$$

To handle the corrupted data, the labeled samples matrix can be rewritten as $X = X_0 + E = VZ + E$, where $X_0$ is the clean labeled samples matrix and $E$ is the error matrix. The optimization problem is rewritten as

$$\min_{Z,E} \|Z\|_* + \lambda_1 \|Z\|_1 + \lambda_2 \|E\|_1,$$
$$s.t. \ X = VZ + E, \ Z_{ii} = 0, \tag{4}$$

where $\lambda_2$ is a scalar parameter.

Two special cases of the LR-S-PFSR model are given below, in which PFSR with sparse constraint or low-rank constraint are called S-PFSR model and LR-PFSR model, respectively.

The S-PFSR model can be expressed as,

$$\min_{Z} \|Z\|_1,$$
$$s.t. \ X = VZ, \ Z_{ii} = 0, \tag{5}$$

The LR-PFSR model can be expressed as,

$$\min_{Z,E} \|Z\|_* + \partial \|E\|_1,$$
$$s.t. \ X = VZ + E, \ Z_{ii} = 0, \tag{6}$$

where $\partial$ is scalar parameter. ADMM [18] can be a solver for the Eqs. (5) and (6).

## 2.3 Optimization based on M-ADMM

Since the objective function and constraint conditions in Eq. (3) are inseparable with respect to low rank and sparse constraints, an auxiliary variable $A$ is introduced. We convert Eq. (4) into an equivalent optimization problem:

$$\min_{Z,E,A} \| Z \|_* + \lambda_1 \| A \|_1 + \lambda_2 \| E \|_1, \qquad (7)$$
$$s.t. \quad X = VZ + E, \ Z - A = 0, \ Z_{ii} = 0, \ A_{ii} = 0.$$

However, Eq. (7) is a multi-block convex minimization problem. The core difficulty lies in the optimization and convergence of the model.

### 2.3.1 Existence of a pair of orthogonal matrices

In order to ensure that the model (7) can be optimized by the M-ADMM [20], we verify the LR-S-PFSR model contains a pair of orthogonal matrices.

For the two constraints in Eq. (7)

$$X = VZ + E, \text{ and } Z - A = 0,$$

which can be converted into a joint constraint

$$\begin{pmatrix} 0 \\ -I \end{pmatrix} A + \begin{pmatrix} V \\ I \end{pmatrix} Z + \begin{pmatrix} I \\ 0 \end{pmatrix} E = \begin{pmatrix} X \\ 0 \end{pmatrix},$$

where $(\cdot)$ before $A$, $Z$ and $E$ represents matrix. The Eq. (7) can be converted into an equivalent optimization problem as

$$\min_{Z,E,A} \| Z \|_* + \lambda_1 \| A \|_1 + \lambda_2 \| E \|_1,$$
$$s.t. \ \begin{pmatrix} 0 \\ -I \end{pmatrix} A + \begin{pmatrix} V \\ I \end{pmatrix} Z + \begin{pmatrix} I \\ 0 \end{pmatrix} E = \begin{pmatrix} X \\ 0 \end{pmatrix}, \ Z_{ii} = 0, \ A_{ii} = 0, \qquad (8)$$

where $\begin{pmatrix} 0 \\ -I \end{pmatrix}^T \times \begin{pmatrix} I \\ 0 \end{pmatrix} = 0$. Thus, the objective function Eq. (8) contains a pair of orthogonal matrices. Similar to [20], Eq. (8) can be solved by the M-ADMM and give its convergence bound. According to [19,20], the corresponding convergence can be guaranteed.

### 2.3.2 Construction of two projection operators

Next, two projection operators are constructed to guarantee that $Z_{ii}$ and $A_{ii}$ in the coefficient matrix of pseudo-full-space are 0.

For the restriction $Z_{ii} = 0$, $Z \in R^{(m+s_c) \times s_c}$, let $\mho = R^{(m+s_c) \times s_c}$ is a closed convex set. Doing a projection operator projects $Z$ into $\mho$,

$$P_\mho(Z) \in \{ Z \in R^{(m+s_c) \times s_c} \mid Z_{ii} = 0, \ i = 1, \cdots, s_c \}.$$

That is,

$$(P_\mho(Z))_{qi} = \begin{cases} Z_{qi}, & if \ q \neq i, \\ 0, & if \ q = i, \end{cases} \quad Z_{qi} \in Z, \tag{9a}$$

where $(P_\mho(Z))_{qi}$ is a component of $P_\mho(Z)$. Then the restriction $Z_{ii} = 0$ is converted to $Z = P_\mho(Z)$.

Similar to $Z_{ii} = 0$, the restriction $A_{ii} = 0$, $A \in R^{(m+s_c) \times s_c}$, can be projected into $\mho$ by doing a similar projection operator,

$$P_\mho(A) \in \{A \in R^{(m+s_c) \times s_c} \mid A_{ii} = 0, i = 1, \cdots, s_c\}.$$

That is,

$$(P_\mho(A))_{qi} = \begin{cases} A_{qi}, & if \ q \neq i, \\ 0, & if \ q = i, \end{cases} \quad A_{qi} \in A, \tag{9b}$$

where $(P_\mho(A))_{qi}$ is a component of $P_\mho(A)$. Then the restriction $A_{ii} = 0$ is converted to $A \in P_\mho(A)$.

### 2.3.3 Optimization

Based on the existence of a pair of orthogonal matrices and the constructions of two projection operators for $Z_{ii} = 0$ and $A_{ii} = 0$, the optimization problem (8) can be solved by the M-ADMM.

The augmented Lagrangian function of Eq. (8) is

$$L_\mu(Z, A, E; T) = \|Z\|_* + \lambda_1 \|A\|_1 + \lambda_2 \|E\|_1 - \left\langle T, \begin{pmatrix} V \\ I \end{pmatrix} Z + \begin{pmatrix} 0 \\ -I \end{pmatrix} A + \begin{pmatrix} I \\ 0 \end{pmatrix} E - \begin{pmatrix} X \\ 0 \end{pmatrix} \right\rangle$$

$$+ \frac{\mu}{2} \left\| \begin{pmatrix} 0 \\ -I \end{pmatrix} A + \begin{pmatrix} V \\ I \end{pmatrix} Z + \begin{pmatrix} I \\ 0 \end{pmatrix} E - \begin{pmatrix} X \\ 0 \end{pmatrix} \right\|_F^2,$$

where $T$ is a Lagrange multiplier, and $\mu > 0$ is a penalty parameter. In each iteration, the updating rules are given as

$$\begin{cases} Z^{k+1} = \arg\min_{Z} \|Z\|_* - \left\langle T^k, \begin{pmatrix} V \\ I \end{pmatrix} Z \right\rangle + \frac{\mu}{2} \left\| \begin{pmatrix} V \\ I \end{pmatrix} Z + \begin{pmatrix} 0 \\ -I \end{pmatrix} A^k + \begin{pmatrix} I \\ 0 \end{pmatrix} E^k - \begin{pmatrix} X \\ 0 \end{pmatrix} \right\|_F^2, \\ (A^{k+1}, E^{k+1}) = \arg\min_{A,E} \lambda_1 \|A\|_1 + \lambda_2 \|E\|_1 - \left\langle T^k, \begin{pmatrix} 0 \\ -I \end{pmatrix} A + \begin{pmatrix} I \\ 0 \end{pmatrix} E \right\rangle \\ \qquad + \frac{\mu}{2} \left\| \begin{pmatrix} V \\ I \end{pmatrix} Z^{k+1} + \begin{pmatrix} 0 \\ -I \end{pmatrix} A + \begin{pmatrix} I \\ 0 \end{pmatrix} E - \begin{pmatrix} X \\ 0 \end{pmatrix} \right\|_F^2, \\ T^{k+1} = T^k - \mu \left( \begin{pmatrix} V \\ I \end{pmatrix} Z^{k+1} + \begin{pmatrix} 0 \\ -I \end{pmatrix} A^{k+1} + \begin{pmatrix} I \\ 0 \end{pmatrix} E^{k+1} - \begin{pmatrix} X \\ 0 \end{pmatrix} \right). \end{cases} \quad (10)$$

Because of $\begin{pmatrix} 0 \\ -I \end{pmatrix}^T \times \begin{pmatrix} I \\ 0 \end{pmatrix} = 0$, $A^{k+1}$ and $E^{k+1}$ are separable. Eq. (10) can be extended to

$$\begin{cases} Z^{k+1} = \arg\min_{Z} L_\mu(Z, A^k, E^k; T^k), \\ A^{k+1} = \arg\min_{A} L_\mu(Z^{k+1}, A, E^k; T^k), \\ E^{k+1} = \arg\min_{E} L_\mu(Z^{k+1}, A^{k+1}, E; T^k), \\ T^{k+1} = T^k - \mu \left( \begin{pmatrix} 0 \\ -I \end{pmatrix} A^{k+1} + \begin{pmatrix} V \\ I \end{pmatrix} Z^{k+1} + \begin{pmatrix} I \\ 0 \end{pmatrix} E^{k+1} - \begin{pmatrix} X \\ 0 \end{pmatrix} \right). \end{cases} \quad (11)$$

Please see Appendix A for details of the optimization process of solving LR-S-PFSR model with M-ADMM, where the major computational complexity is majorly dominated by the SVD of matrix. So the complexity of LR-S-PFSR is $O\left(k(s_c + m)s_c^2\right)$, where $k$ is the iteration number.

The corresponding convergence analysis is conducted by giving three lemmas and a convergence theorem. Please see Appendix B for the detail convergence analysis. In Subsection 5.3, experiments further verify the convergence of the LR-S-PFSR model. Moreover, similar to [20], the convergence speed of Eq. (7) can be characterized by the convergence bound theorem in Appendix B.

## 3 Performance analysis

The performance analysis of the proposed LR-S-PFSRC model is discussed, which includes representation capability and classification stability. A category concentration index and a relative stability index are constructed to quantify the representation capability and classification stability, respectively.

**3.1 Representation capability**

Similar to PFSR [12], a statistical index, category concentration index (CCI), is introduced to measure the representation capability of coefficient vectors. Suppose $\alpha_l \in R^{s_c}$ is a coefficient vector about labeled samples, $\delta_i$ is a characteristic function, $\delta_i(\alpha_l) \in R^{s_c}$ is a vector whose nonzero entries are the entries that are associated with the $i$-th category in $\alpha_l$. The CCI of $\alpha_l$ is defined as

$$CCI(\alpha_l) = \frac{\max_i \|\delta_i(\alpha_l)\|_1}{\|\alpha_l\|_1}.$$

The larger the CCI is, the better the representation is. Owing to the difference from the PFSR model and the LR-S-PFSR model, the representation coefficient matrices of the two models are different. And then the CCI of the LR-S-PFSRC model is different from that of PFSR model. The detailed examples and experiment results will be given in subsection 5.4.2.

**3.2 Classification Stability**

In order to measure classification stability, a statistical index, relative stability index (RSI), is constructed. In addition, the stability analysis of the LR-S-PFSRC model and corresponding classification stability theorem, please see Appendix B.

In order to measure classification stability, the relative stability index corresponding to $a_{ij}$ can be constructed. The relative stability index indicates the reduction of classification accuracy when the optimal point turns to the measured point in different models of a certain environment. The relative stability index reveals location of classification accuracy for measured point relative to the optimal classification accuracy.

**Definition 3.1 (Relative Stability Index, RSI）:**

$$RSI_{ij} = \frac{\max(A) - a_{ij}}{a_{ij}}, \tag{12}$$

where $a_{ij}$ is a classification accuracy of $j$ labeled samples per category corresponding to the $i$-th model. The recognition results matrix of all models in different cases can be represented as $A = [a_1, \cdots, a_i, \cdots]$, where $a_i = [a_{i1}, \cdots, a_{ij}, \cdots]$. $RSI_{ij}$ represents the relative stability index of the $i$-th model on $j$ labeled samples per category. According to Eq. (12), one can see $RSI_{ij} \geq 0$. The smaller the $RSI_{ij}$ is, the more stable the model is. Conversely, the larger the $RSI_{ij}$ is, the worse the stability of model is.

## 4 Face recognition algorithm based on the LR-S-PFSRC model

For completing the classification process, similar to PFSRC [12], the category contribution rate (CCR) is constructed as classification criterion, which indicates the relevance between a unlabeled sample and a category. The larger the CCR is, the stronger the corresponding relevance is.

For a unlabeled sample $y_l$, the contribution rate $C_{j,l}$ of $y_l$ in the $j$-th category can be calculated by,

$$C_{j,l} = \frac{1}{n_j} \sum_i \frac{\delta_j\left(\{\beta_{i,l}\}_{i=1,\cdots,s_c}\right)}{\left\|\{\beta_{i,l}\}_{i=1,\cdots,s_c}\right\|_1}, \quad (13)$$

where $\beta_{i,l}$ is a coefficients vector corresponding to the $i$-th $y_l$, and $\delta_j(\cdot)$ is a vector whose entries are zero except those associated with the $j$-th category. Note that $j = 1, \cdots, c$, $l = 1, \cdots, m$, $i = 1, \cdots, s_c$, and $n_j$ denotes the number of $j$-th category labeled samples. And then the CCR matrix $[C_{j,l}]$ for all unlabeled samples can be calculated. The unlabeled sample $y_l$ is classified into the category with the maximal CCR $[C_{j,l}]$.

What is more, the CCR matrix $[C_{j,l}]$ can be seen as membership degree matrix of all categories for each unlabeled sample. Owing to the difference of PFSR and LR-S-PFSR, the coefficient matrices are different, and then the CCR of LR-S-PFSRC model is different to that of the PFSR.

The LR-S-PFSRC model for face recognition is outlined in Algorithm 1. Fig. 2 shows the framework of face recognition based on LR-S-PFSRC.

**Algorithm 1**. The LR-S-PFSRC model for face recognition.

**Input:** Labeled data matrix $X = [X_1, X_2, \cdots, X_c] \in R^{d \times s_c}$ associated with label set $\Gamma = [l_1, l_2, \cdots, l_c] \in R^{s_c}$, unlabeled data matrix $Y = [y_1, y_2, \cdots, y_m] \in R^{d \times m}$, parameter $\lambda_1, \lambda_2$.

**Output:** the label of $y_l$.

1. Calculate the eigenface matrix $U$ of the face image using the labeled data.
2. Calculate labeled data and unlabeled data after extracting features: $X = U^T X$, $Y = U^T Y$.
3. Normalize the columns of $X$ and $Y$ to have unit $l_2$-norm.
4. Initialize: $Z^0 = A^0 = E^0 = T^0 = 0$, $k = 1$.
5. While not converge do

6. Update $Z$: $Z^{k+1} = D_{\eta_V^{-1}}(Z^k - \dfrac{B^k}{\eta_V})$ and $(P_\mho(Z^{k+1}))_{qi} = \begin{cases} Z_{qi}^{k+1}, & \text{if } q \neq i, \\ 0, & \text{if } q = i, \end{cases}$ $Z_{qi}^{k+1} \in Z^{k+1}$;

7. Update $A$: $A^{k+1} = S_{\frac{\lambda_2}{\mu_2}}\left(Z^{k+1} - \dfrac{T_2^k}{\mu_2}\right)$ and $(P_\mho(A^{k+1}))_{qi} = \begin{cases} A_{qi}^{k+1}, & \text{if } q \neq i, \\ 0, & \text{if } q = i, \end{cases}$ $A_{qi}^{k+1} \in A^{k+1}$;

8. Update $E$: $E^{k+1} = S_{\frac{\lambda_2}{\mu_1}}(X - VZ^{k+1} + \dfrac{T_1^k}{\mu_1})$;

9. Update $T$: $\begin{cases} T_1^{k+1} = T_1^k - \mu_1\left(VZ^{k+1} + E^{k+1} - X\right); \\ T_2^{k+1} = T_2^k - \mu_2\left(Z^{k+1} - A^{k+1}\right); \\ T = (T_1, T_2)^T; \end{cases}$

10. $k = k+1$;
11. End while

12. Calculate the category contribution matrix $C_{j,l} = \dfrac{1}{n_j} \sum_i \dfrac{\delta_j\left(\{\beta_{i,l}\}_{i=1,\cdots,s_c}\right)}{\left\|\{\beta_{i,l}\}_{i=1,\cdots,s_c}\right\|_1}$.

13. $identity(y_l) = \underset{j \in \{1,2,\cdots,c\}}{\arg\max}(C_{j,l})$.

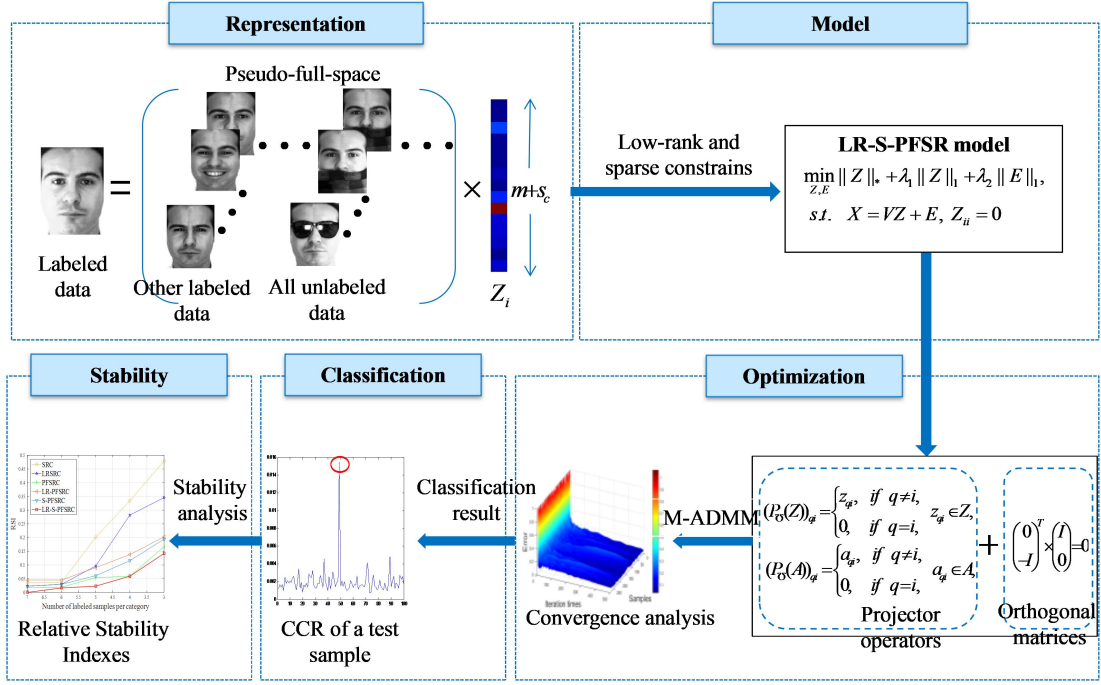

Fig. 2. The framework of LR-S-PFSRC for robust face recognition.

## 5 Experiments and discussions

Experiments mainly discuss parameter analysis, convergence, representation performance, classification performance and stability. Since PFSRC [12] has been verified to be superior to some classical classifiers, there are comparison methods including SRC [1], SRC with adding unlabeled samples (SRC+test), low-rank sparse representation-based classification (LRSRC) [6], LRSRC with adding unlabeled samples (LRSRC+test), pseudo-full-space representation based classification (PFSRC) [12], PFSRC with low-rank constraint (LR-PFSRC) and PFSRC with sparse constraint (S-PFSRC). We select category concentration index (CCI), classification accuracy and relative stability index (RSI) as evaluation indicators. All the experiments are carried out using MATLAB R2016a on a 3.30 GHz machine with 4.00GB RAM.

### 5.1 Datasets

As stated in [1], there are two assumptions in standard SRC: (1) experiments are confined to human frontal face recognition, which allows small variations in pose and displacement. (2) detection, cropping, and normalization of the images have been performed prior to applying the corresponding algorithms. Similar to SRC [1], this paper also considers the data obtained under these conditions. Three standard face datasets are adopted, including AR dataset [21], Extended Yale B dataset [22] and CMU Multi-PIE dataset [23].

AR dataset contains over 4000 face images of 126 subjects with 26 images per subject. The 26 images from each subject can be divided into two sections. Each section contains 1 natural image, 3 images with expression, 3 images with illumination, 3 images with sunglasses and 3 images with scarf. As done in [24], we choose a subset containing 50 male subjects and 50 female subjects in the experiments. Fig. 3 (a) shows the whole images of one subject in AR dataset.

Extended Yale B dataset contains 38 subjects with about 2414 frontal face images for all subjects. Each individual has about 64 images taken under various laboratory-controlled lighting conditions. In our experiments, each image is manually cropped and resized to $32 \times 32$ pixels. Fig. 3 (b) shows several example images of one subject in Extended Yale B dataset.

CMU Multi-PIE is a dataset with 68 subjects and 41368 images. These images of each subject are obtained by 13 different poses, 43 different illuminations and 4 different expressions. We use a near frontal pose subset for experiments, namely C07, which contains 1629 images of 68 individuals, and each individual has about 24 images. In our experiments, all images are manually cropped and resized to be $64 \times 64$ pixels. Some example images of one subject are shown in Fig. 3 (c).

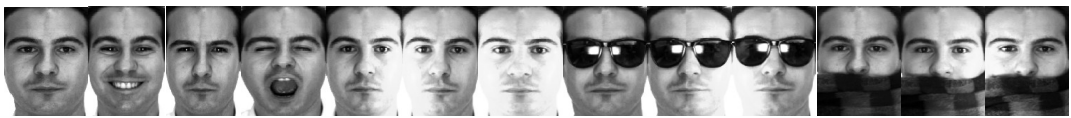

(a)

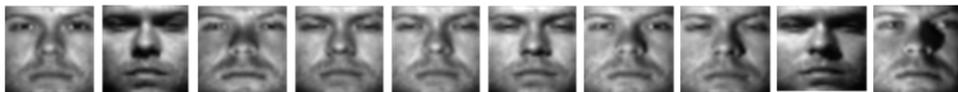

(b)

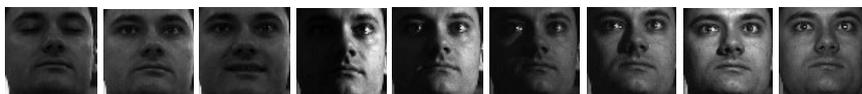

(c)

Fig. 3. Some examples. (a), (b) and (c) are AR, Extended Yale B and CMU Multi-PIE datasets.

## 5.2 Parameter analysis

Penalty parameters $\mu_1$ and $\mu_2$ affect the convergence and classification accuracy of the proposed model. In this paper, by referring to [25], $\mu_1$ ranges from 0.3 to 0.7 and $\mu_2$ ranges from 0.02 to 0.06. The parameters are selected heuristically according to classification accuracy. Note that samples are extracted the eigenfaces of 300 dimensions based on principal component analysis, and the number of iteration steps is set to 500.

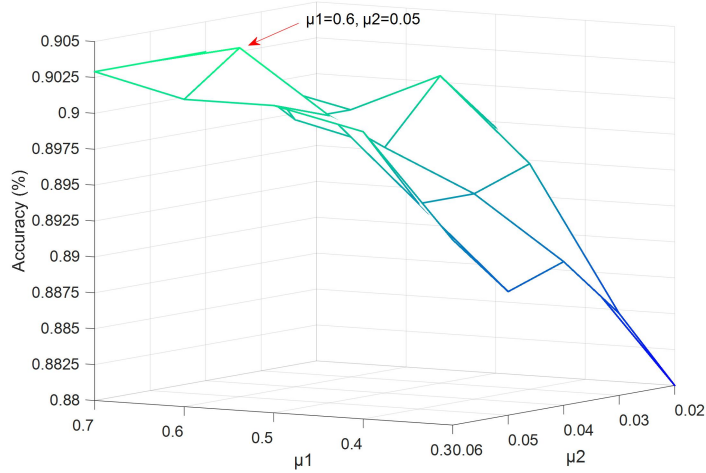

Fig. 4. Accuracies based on LR-S-PFSRC with different $\mu_1, \mu_2$.

The accuracies of different penalty parameters $\mu_1$ and $\mu_2$ are shown in Fig. 4, where it shows that the parameters $\mu_1$ and $\mu_2$ have a great influence on the classification results, and the maximum is reached when $\mu_1 = 0.6$ and $\mu_2 = 0.05$. Hence, the subsequent experiments in this paper all use these parameter settings.

## 5.3 Convergence analysis

In subsections 2.3, the LR-S-PFSR has been optimized with the M-ADMM and the corresponding convergence is proved. Here, the convergence of LR-S-PFSR model is verified on AR, Extended Yale B and CMU Multi-PIE datasets. Fig. 5 shows two types of iterative error graphs of LR-PFSR, S-PFSR and LR-S-PFSR models on AR dataset. It can be seen that the convergence error of LR-S-PFSR is the smallest, and the convergence speed is the fastest.

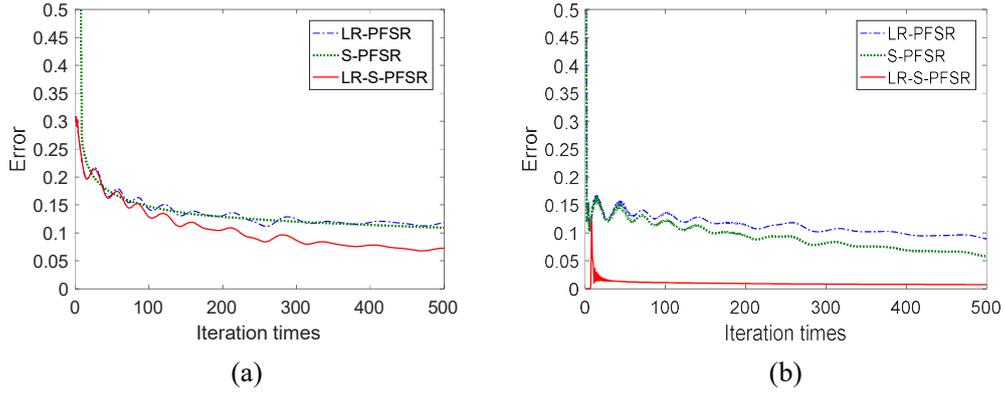

Fig. 5. Comparison of iteration error. (a) The error between iterative solution and real solution, (b) the error between the adjacent iterations.

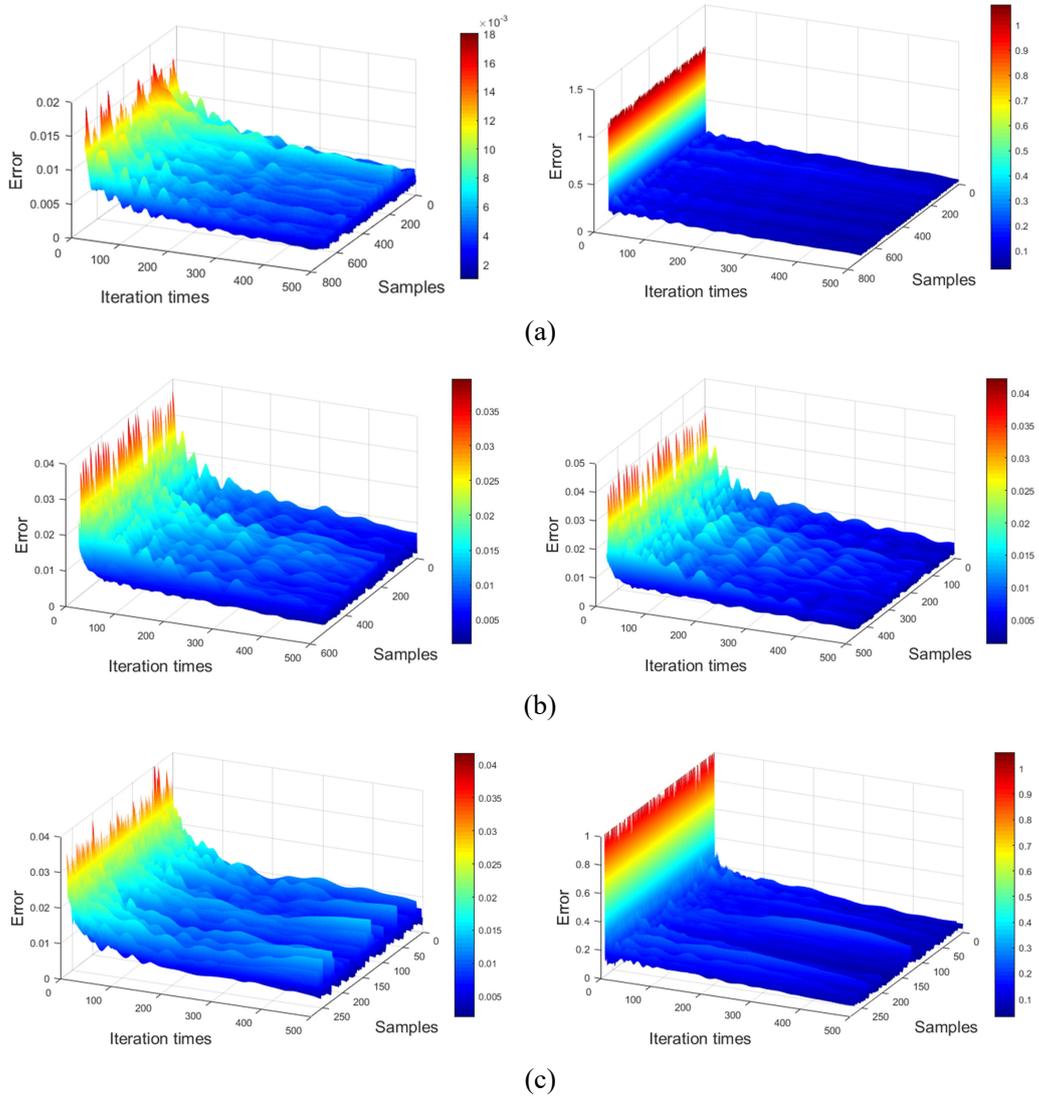

Fig. 6. The iteration error of the LR-S-PFSR model between iterative solution and real solution (left column), the iteration error between the adjacent iterations (right column). (a) AR dataset, (b) Extended Yale B dataset and (c) CMU Multi-PIE dataset.

Fig. 6 shows the iterative error trend for LR-S-PFSRC model on three datasets, where the red to blue indicates the iteration error is from high to low. The left column are the iteration errors between iterative solution and real solution, which indicates the iterative solution gradually approaches the real solution and the constraints in model are gradually satisfied. The right column are the iteration errors between the adjacent iterations. From Fig. 6, one can see that the overall trend of errors declines and tends to zero, which demonstrates the convergence.

### 5.4 Representation performance analysis

In this subsection, the representation performance analysis of LR-S-PFSR model contains the role of unlabeled samples and comparison are discussed.

### 5.4.1 The role of unlabeled samples in representation

For accessing the role of unlabeled samples in LR-S-PFSR, experiments are conducted on AR, Extended Yale B and CMU Multi-PIE datasets. There are comparison methods including SRC, SRC+test, LRSRC, LRSRC+test, PFSRC and LR-S-PFSRC. Based on the characteristics of datasets, the settings for different datasets are as follows. For the AR dataset, the labeled samples per category are fixed to 4 and the number of unlabeled samples per category are gradually increased from 2，4，5 to 6. The dimensions of eigenface feature are all set to 150. For the Extend Yale B dataset, the labeled samples per category are fixed to 4, and the number of unlabeled samples per category are gradually increased from 2，4，8 to 12. Here, the dimensions of eigenface feature are set to 150. For the CMU Multi-PIE dataset, the labeled samples per category are fixed to 4, and the number of unlabeled samples per category are gradually increased from 2，4，6 to 8. The dimensions of eigenface feature are set to 200. One can see that no matter which dataset, the number of unlabeled samples is divided into three cases: smaller than the number of labeled samples, equal to the number of labeled samples and more than the number of labeled samples. Overall, the number of unlabeled samples increases in turn. Table 1 gives the accuracies of different classification methods, when the number of labeled samples is fixed and the number of unlabeled samples is sequentially increased. From Table 1, the accuracies of SRC+test are generally lower than other methods, and LR-S-PFSRC achieves the highest result.

Table 1. Classification accuracies (%) for different numbers of unlabeled samples on three datasets.

| Datasets | Methods | number of unlabeled samples per category | | | |
|---|---|---|---|---|---|
| | | 2 | 4 | 5 | 6 |
| AR dataset | SRC | 87.50 | 82.00 | 78.20 | 74.00 |
| | SRC+test | 68.00 | 69.00 | 65.20 | 57.33 |
| | LRSRC | 89.00 | 87.25 | 82.60 | 74.50 |
| | LRSRC+test | 92.50 | 88.00 | 87.40 | 80.00 |
| | PFSRC | 91.50 | 88.75 | 86.40 | 84.17 |
| | **LR-S-PFSRC** | **94.00** | **90.50** | **88.80** | **88.00** |
| | | 2 | 4 | 8 | 12 |
| Extended Yale B dataset | SRC | 100 | 98.03 | 95.07 | 89.91 |
| | SRC+test | 96.05 | 92.76 | 59.21 | 69.08 |
| | LRSRC | 100 | 97.37 | 95.39 | 89.91 |
| | LRSRC+test | 100 | 98.68 | 95.39 | 92.11 |
| | PFSRC | 100 | 100 | 98.68 | 95.18 |
| | **LR-S-PFSRC** | **100** | **99.34** | **98.68** | **95.61** |
| | | 2 | 4 | 6 | 8 |
| CMU Multi-PIE dataset | SRC | 57.35 | 47.79 | 46.32 | 42.10 |
| | SRC+test | 44.85 | 25.00 | 20.83 | 15.99 |
| | LRSRC | 62.50 | 53.68 | 50.25 | 47.98 |
| | LRSRC+test | 85.29 | 77.57 | 77.45 | 75.92 |
| | PFSRC | 99.26 | 91.91 | 88.48 | 84.38 |
| | **LR-S-PFSRC** | **99.26** | **93.01** | **89.22** | **86.76** |

As can be seen from Table 1, with the increase of the number of unlabeled samples, LR-S-PFSRC always gives the highest classification accuracy in all methods. Compared with SRC and PFSRC, the classification accuracy of PFSRC is higher. In particular, when there are few labeled samples, the different projection modes make the sensitivity of the inverse projection representation to the number of labeled samples is less than that of the positive projection representation. In addition, when unlabeled samples is added to the representation space of SRC, that is representation spaces of PFSRC and SRC+test are same. It is found that the unlabeled samples of SRC+test representation space do not play an auxiliary role in classification, but lower the classification accuracy. The classification accuracy of SRC+test is generally decreased, which is due to the fact that the proportion of labeled samples in the representation space decreases because the unlabeled samples have no labels.

Without causing confusion, LRSRC with adding unlabeled samples, namely LRSRC+test, is to add unlabeled samples in the representation space of LRSRC. Compared with LRSRC and LRSRC+test, the classification accuracy of LRSRC+test is generally increased, which is due to the fact that LRSRC model contains low-rank constraint which explores the structures information of data, regardless of whether the data contains labels or not. Therefore, due to the addition of unlabeled samples, LRSRC+test enhances the representation capability of model by mining the structures information and complementary information of data. So, the classification accuracy of LRSRC+test is higher than that of LRSRC. That is to say, SRC+test can not improve the classification accuracy because SRC model has no low rank constraint. Even if the unlabeled samples is added to the representation space of SRC, it can not make use of the information contained in the unlabeled samples, which means the intrinsic structure of data can not be explored. On the contrary, it can lower the proportion of labeled samples. In addition, the sparse constraint and the low-rank constraint are added on the basis of the PFSRC to obtain the LR-S-PFSRC. Comparing LRSRC+test and LR-S-PFSRC, their representation spaces are same, but the classification accuracy of LR-S-PFSRC is higher than that of LRSRC+test. The reason is that LR-S-PFSRC is inverse projection representation, while LRSRC+test is positive projection representation. The dependence of LR-S-PFSRC on labeled samples is less than that of LRSRC+test for few labeled samples problem.

### 5.4.2 Representation capability

To verify that LR-S-PFSR has stronger representation capability than PFSR, the representation capability can be demonstrated by comparing the CCI results.

Fig. 7 gives the CCI values between PFSR (blue lines) and LR-S-PFSR (red lines) on AR dataset and Extended Yale B dataset. The bigger the CCI is, the better the representation is. One can see that for all samples, $CCI_{LR-S-PFSR} > CCI_{PFSR}$ is almost always satisfied. Fig.7 shows that LR-S-PFSR has better representation performance than PFSR. The main reason is the coefficients obtained by LR-S-PFSR are more consistent with the intrinsic characteristics of data than PFSR.

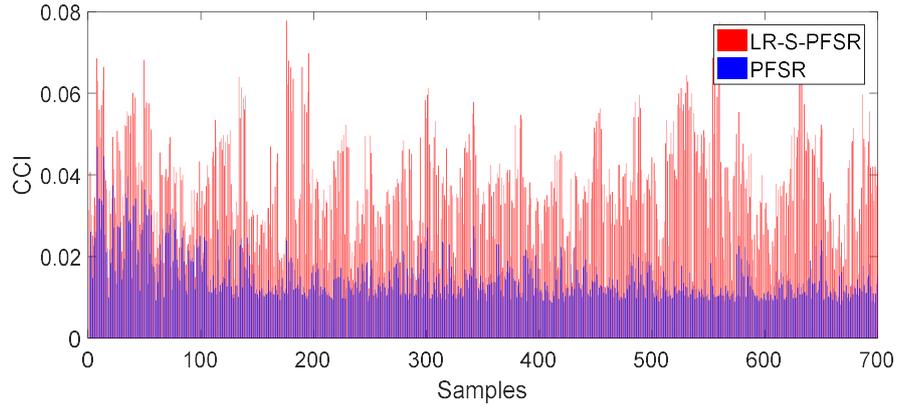

(a)

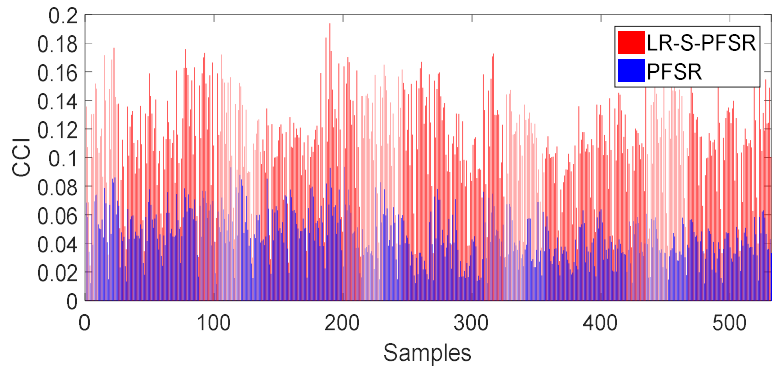

(b)

Fig. 7. Comparison of CCI values. (a) AR dataset, (b) Extended Yale B dataset.

### 5.5 Performance of category contribution rate

In addition to the representation method, the classification criterion has a significant impact on SRC-based methods. To demonstrate the classification performance of the category contribution rate (CCR), some unlabeled samples are randomly selected from AR dataset as examples. Fig. 8 gives the CCR results of these unlabeled samples. As the Fig. 8 shows, it is sparse that the sample contributes to the categories. That is, there is only one peak (circled in red) in every subfigure, the CCR of one category is far greater than other categories, and then the unlabeled sample should belong to the corresponding category.

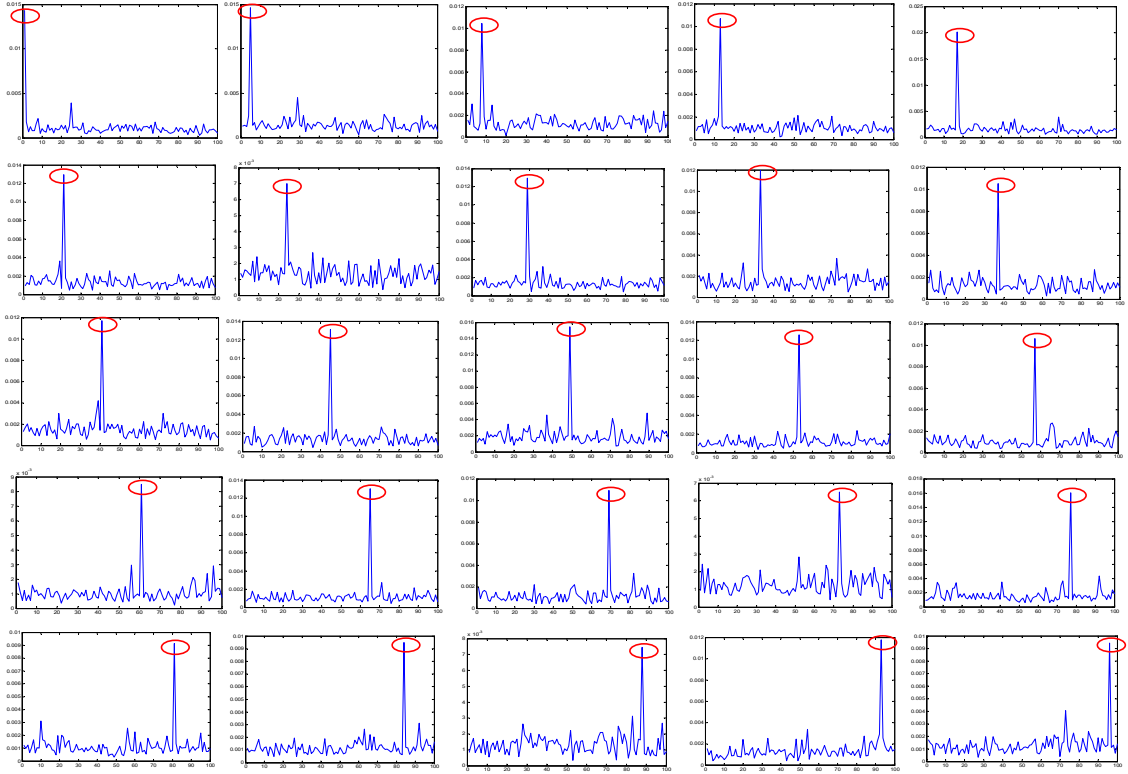

Fig. 8. The CCRs of some unlabeled samples versus all categories on AR dataset.

**5.6 Stability analysis**

The stability of the proposed LR-S-PFSRC model is verified by considering the varieties of expression, illumination, different occlusions and proportion imbalance of samples. For the proportion imbalance between labeled samples and unlabeled samples, there are two cases: one is that the number of unlabeled samples is fixed, and the number of labeled samples is decreased in turn. The other is that the number of labeled samples in each category is fixed, and the number of tests samples is increased in turn.

**5.6.1 Stability to proportion imbalance of samples with decreasing labeled samples**

The following experiments will verify the stability of the proposed methods when the labeled samples are gradually decreased. The regularization parameters are set as $\lambda_1 = 10$ and $\lambda_2 = 0.09$ by experiments and experience. On the AR dataset, for each subject, we select 7 different light images as labeled samples, leave 7 different light images as unlabeled samples, and gradually reduce the number of labeled samples per category from 7 to 3. The feature dimension is set to 200. On the Extend Yale B dataset, the number of labeled samples per category are selected as $l(=32,28,24,20,16,12,8,4)$ and 24 samples per category are selected as unlabeled samples among the remaining samples. The feature dimension is set to 150. On the CMU Multi-PIE dataset, the C07 dataset is used for experiments. The number of labeled samples per category decreases gradually from 8 to 2 and to select 10 samples per category as unlabeled samples. Fig. 9 shows classification accuracies (the first row), RSIs (the middle row) and rose figures of RSI (the bottom row) with decreasing number of labeled samples on AR, Extended Yale B and CMU Multi-PIE datasets.

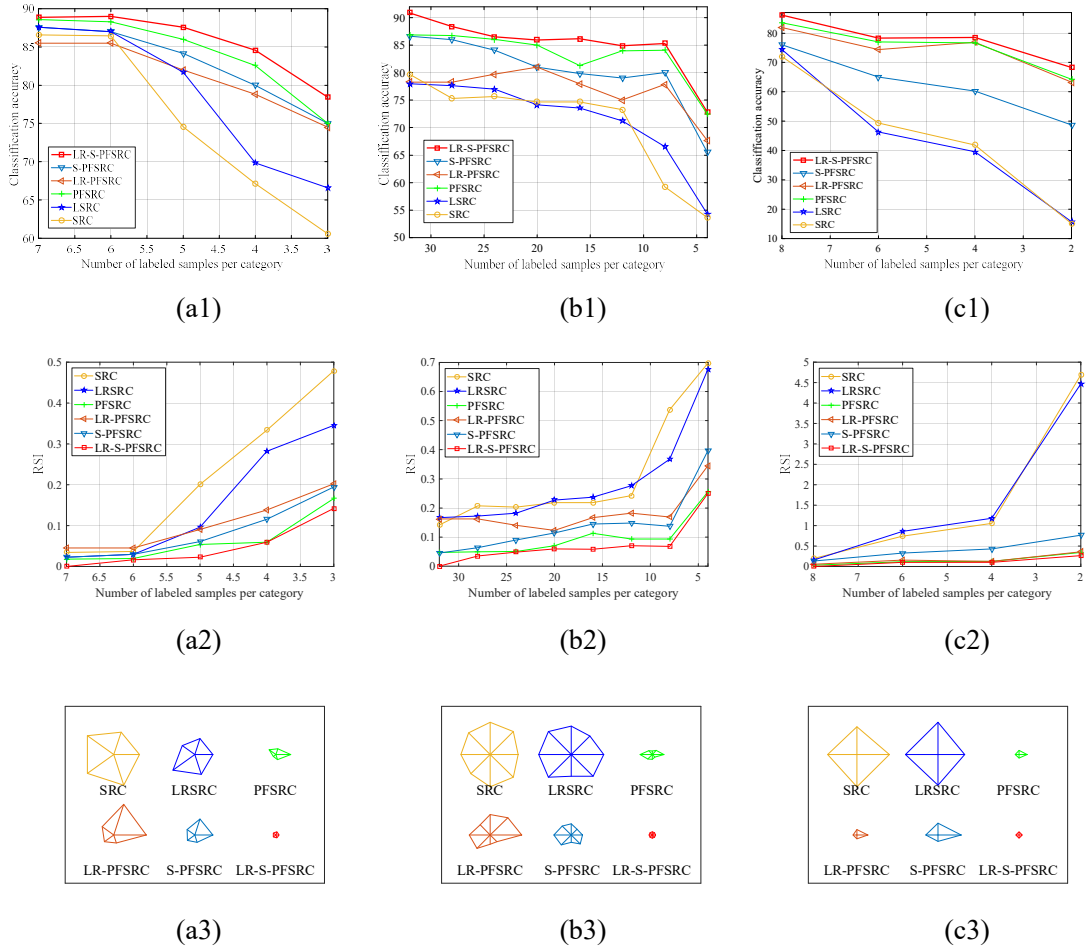

Fig. 9. Comparison of classification accuracies and stability with decreasing number of labeled samples on (a1)-(a3) AR, (b1)-(b3) Extended Yale B and (c1)-(c3) CMU Multi-PIE datasets. From the first row to the third row are classification accuracies, RSIs and rose figure of RSI, respectively.

The rose figures of RSI in Fig. 9 measures the classification stability of model. For a rose figure, the distance from each endpoint to the center corresponds to each RSI value. The shorter the distance is, the smaller the RSI value is. The smaller the RSI value is, the more concentrated the rose figure is, which means the more stable the model is. We take the rose figure of RSI for SRC on AR dataset as an example. In Fig. 9(a2), one can see that there are 5 data points on the line figure of RSI for SRC. The RSI value of each data point corresponds to the distance from each endpoint to the center on the rose figure of RSI for SRC in Fig. 9(a3), which is similar to other rose figures. In Figs. 9(a2)-(a3), the line figure of LR-S-PFSRC is the lowest in all methods, and the corresponding rose figure is the most concentrated, which means the LR-S-PFSRC model achieves competitive stability. As one can see from Fig. 9, the LR-S-PFSRC model has the highest classification accuracies, the smallest RSI values, and the most concentrated rose figures in all methods for each dataset. That is the LR-S-PFSRC model has better classification performance and stability than other methods. The accuracies of other methods decrease significantly with the decrease of labeled samples, but our method can still achieve a relatively stable recognition effect. It is clear that the comparison methods rely more on sufficient labeled samples per category while our model is meaningful in solving the few labeled samples problem.

**5.6.2 Stability to proportion imbalance of samples with increasing unlabeled samples**

To further verify the stability of the proposed model, the number of labeled samples per category are fixed and the number of unlabeled samples are gradually increased. For the AR dataset, the labeled samples per category are fixed to 2, 3, 4 and 5 respectively, and the number of unlabeled samples per category are gradually increased from 2 to 7. The dimensions of eigenface feature are all set to 150. For the Extend Yale B dataset, the labeled samples per category are fixed to 4, 6, 8 and 10, and the number of unlabeled samples per category are gradually increased from 4, 8, 12, 16, 20 to 24. Here, when the labeled sample per category is 4, 6, 8 and 10, the dimensions of eigenface feature are set to 150, 200, 200 and 200, respectively. For the PIE datasets, the labeled samples per category are fixed to 2, 4, 6 and 8, and the number of unlabeled samples per category are gradually increased from 2 to 10. Here, when the labeled sample is 2, 4, 6 and 8 per class, the dimensions of eigenface feature are set to 100, 200, 300 and 300, respectively. With increasing the number of unlabeled samples, the result of classification accuracy, RSIs and rose figures of RSI are presented in Fig. 10.

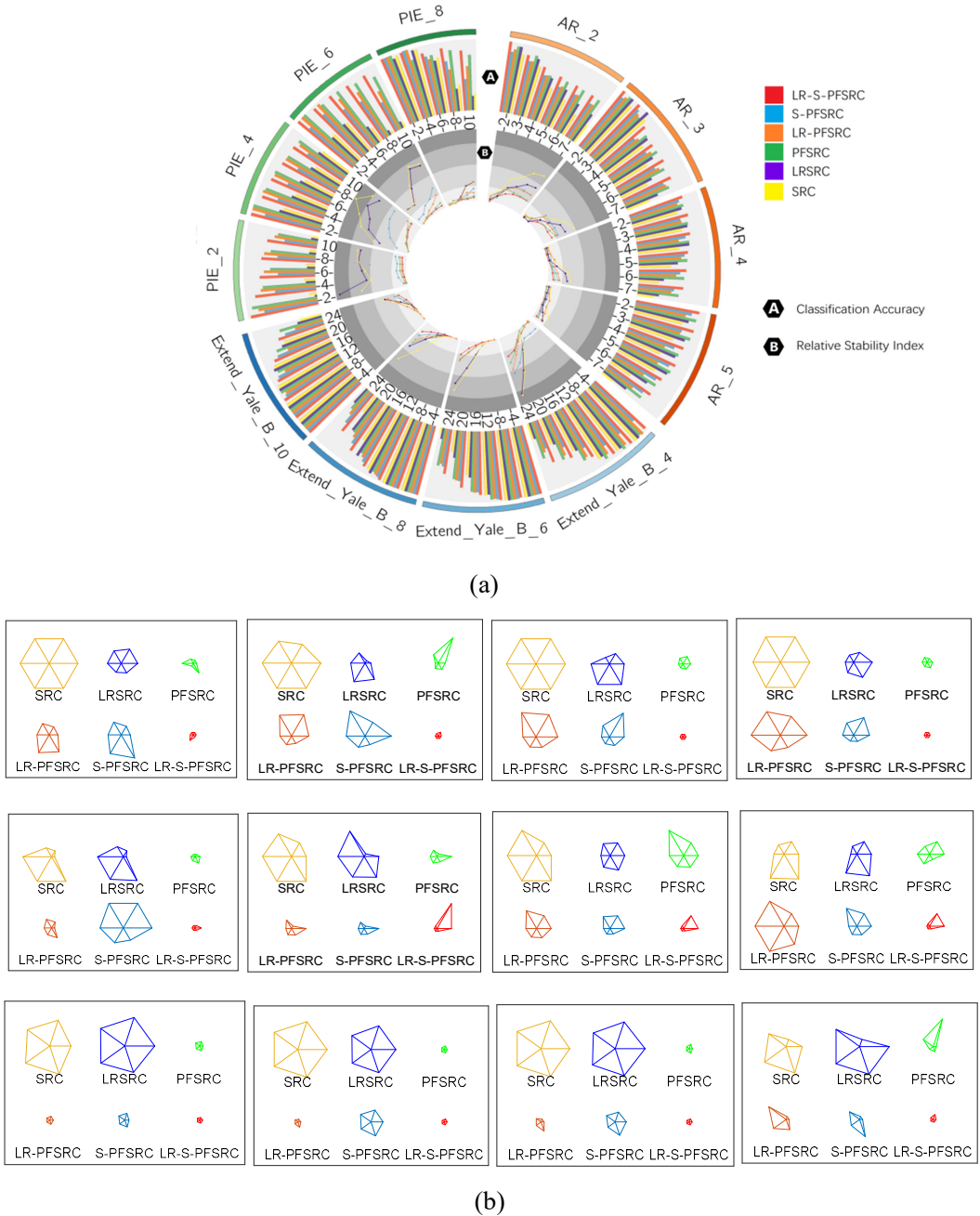

(a)

(b)

Fig. 10. Classification accuracies and stability with increasing number of unlabeled samples. (a) classification accuracies (the outer circle) and RSIs (the inner circle) are respectively ploted on the circular cartogram, (b) the three rows are the rose figures of RSI, where the first row is results of labeled samples set to 2, 3, 4 and 5 from AR dataset; the second row is results of labeled samples set to 4, 6, 8 and 10 from Extended Yale B dataset; the third row is results of labeled samples set to 2, 4, 6 and 8 from CMU Multi-PIE dataset.

Fig. 10(a) is a circos figure which visualizes data in a circular layout. The most important reason why the circular layout has advantages is that it is very attractive. The outer circle of Fig. 10(a) is classification accuracy histogram, where the vertical axis represents the classification accuracy, the horizontal axis represents the number of unlabeled samples per category, and the columns of different colors represent different classification methods. The inner circle of Fig. 10(a) is line chart of RSI, where the vertical axis represents the RSI values, the horizontal axis represents the number of unlabeled samples per category, and the lines of different colors represent different classification methods. As one can see from Fig. 10(a), we can notice the following phenomena: (1) according to the outer circle of Fig. 10(a), the accuracies of all methods decreases gradually as the number of unlabeled samples increases, while the proposed method outperforms others and decreases slowly. (2) from the inner circle of Fig. 10(a), the lower the RSI curve is, the more stable the method is. It is clear that the RSI curve of LR-S-PFSRC is the lowest in all methods, which indicates our method is relatively stable for proportion change between labeled samples and unlabeled samples. Fig. 10(b) exhibits the rose figures of RSI. As can be seen from Fig. 10(b), the rose figure of RSI for the proposed method is more concentrated, which shows its classification stability is more stronger than other methods. Experimental results verify our method achieves competitive classification accuracy and stability when the labeled samples are too few and the proportion of samples is imbalanced.

### 5.6.3 Stability to different occlusions

The efficiency and stability of the proposed model to various occlusions is further verified on AR dataset. 2 natural images per category are used as labeled samples, and the remaining 24 are used as unlabeled samples. There are four types of occlusions in unlabeled samples: expression, illumination, sunglasses and scarf. The dimension of eigenface feature is 150.

Table 2 shows that the last four PFSRC methods are superior to the first two SRC methods, and the LR-S-PFSRC has the highst classification accuracy in all the four types of occlusions. These suggest that: (1) the PFSR strategy is indeed more suitable than the standard sparse representation for the case of few labeled samples; (2) the PFSR strategy has better stability when dealing with various occlusions than the standard sparse representation; (3) the LR-S-PFSRC superiors the other methods because it can fully mining the useful information contained in the existing data and the complementary information between the data.

Table 2. Classification accuracies (%) with different types occlusions on AR dataset

| Methods | Occlusions | | | |
| --- | --- | --- | --- | --- |
| | Expression | Illumination | Sunglasses | Scarf |
| SRC [1] | 90.17 | 55.17 | 26.00 | 22.33 |
| LRSRC [6] | 97.00 | 60.67 | 32.67 | 31.17 |
| PFSRC [12] | 94.67 | 81.17 | 72.50 | 66.33 |
| S-PFSRC | 95.50 | 82.83 | 73.50 | 68.83 |
| LR-PFSRC | 95.67 | 82.50 | 73.00 | 67.00 |
| **LR-S-PFSRC** | **97.33** | **85.67** | **82.50** | **73.33** |

Except for standard sparse representation-based face recognition methods, comparisons are made to other methods on the same dataset, including LBP [26], P-LBP [27], and PCANet [9]. For each subject, we select images with frontal illumination and neural expression as labeled samples on AR dataset. Meanwhile, images with expression changes and illumination variation are viewed as unlabeled samples Table 3 proves that our algorithm still achieves impressive performance with higher accuracies than other methods, especially for PCANet, a simple deep learning baseline for image classification. Compared with PCANet, LR-S-PFSRC also achieves comparable or even slightly better classification performance, but without learning.

Table 3. Classification accuracies (%) with different types occlusions on AR dataset

| | LBP [26] | P-LBP [27] | PCANet-1 [9] | PCANet-2 [9] | **LR-S-PFSRC** |
| --- | --- | --- | --- | --- | --- |
| Expression | 81.33 | 80.33 | 85.67 | 85.00 | **86.33** |
| Illumination | 93.83 | 97.50 | 98.00 | 99.50 | **100** |

## 6 Conclusions

In this paper, we focus on obtaining the sparsest and lowest-rank representation via simple linear regression strategy and without learning. It is worth noting that the proposed LR-S-PFSRC model is effective and stable for few labeled samples and proportion imbalance of samples problem. In theory, we discuss the optimization and analyze the corresponding convergence, representation capability and classification stability. In addition, RSI is constructed to measure the stability of model.

There remains some interesting problems to be solved. How to exploit the iteratively reweighted least squares [28] without introducing auxiliary variables and computing singular value decomposition to solve low rank minimization problems. In addition, compared with the sparse solutions of the $l_1$ regularization, Xu et al. have show that the $l_{1/2}$ regularization always yields the better sparse solution [29]. On the other hand, the rank function can also be further relaxed. Therefore, we are considering to construct an inverse projection sparse representation-based classification framework with more relaxed regularizations, which has a wider range of applicability. SRC methods may be more focused on a controllable experimental environment. As for the actual IoT environment, deep learning approaches have obvious advantages. It might be interesting and meaningful to complement each other.

**Acknowledgements**

The authors would like to thank the anonymous reviewers and the associate editor for their valuable comments and thoughtful suggestions which improved the quality of the presented work. We also thank Prof. Bingsheng He for helpful optimizations suggestion. This work was supported in part by National Natural Science of China (Grants No. 41771375), Nature Science of Henan Province (Grant No. 202102310087) and Open Fund of Key Laboratory of Intelligent Perception and Image Understanding of Ministry of Education, Xidian University (Grant No. IPIU2019010)

The authors have declared that no conflict of interest exists.

**Appendix A**

The optimization process of LR-S-FSRC model. The augmented Lagrangian function of Eq. (8) as,

$$L_\mu(Z, A, E; T) = \|Z\|_* + \lambda_1 \|A\|_1 + \lambda_2 \|E\|_1 - \left\langle T, \begin{pmatrix} V \\ I \end{pmatrix} Z + \begin{pmatrix} 0 \\ -I \end{pmatrix} A + \begin{pmatrix} I \\ 0 \end{pmatrix} E - \begin{pmatrix} X \\ 0 \end{pmatrix} \right\rangle$$
$$+ \frac{\mu}{2} \left\| \begin{pmatrix} 0 \\ -I \end{pmatrix} A + \begin{pmatrix} V \\ I \end{pmatrix} Z + \begin{pmatrix} I \\ 0 \end{pmatrix} E - \begin{pmatrix} X \\ 0 \end{pmatrix} \right\|_F^2,$$

where $T$ is a Lagrange multiplier, and $\mu > 0$ is a penalty parameter. In each iteration, the updating rules are given as

$$\begin{cases} Z^{k+1} = \arg\min_Z \|Z\|_* - \left\langle T^k, \begin{pmatrix} V \\ I \end{pmatrix} Z \right\rangle + \frac{\mu}{2} \left\| \begin{pmatrix} V \\ I \end{pmatrix} Z + \begin{pmatrix} 0 \\ -I \end{pmatrix} A^k + \begin{pmatrix} I \\ 0 \end{pmatrix} E^k - \begin{pmatrix} X \\ 0 \end{pmatrix} \right\|_F^2, \\ (A^{k+1}, E^{k+1}) = \arg\min_{A,E} \lambda_1 \|A\|_1 + \lambda_2 \|E\|_1 - \left\langle T^k, \begin{pmatrix} 0 \\ -I \end{pmatrix} A + \begin{pmatrix} I \\ 0 \end{pmatrix} E \right\rangle \\ \qquad + \frac{\mu}{2} \left\| \begin{pmatrix} V \\ I \end{pmatrix} Z^{k+1} + \begin{pmatrix} 0 \\ -I \end{pmatrix} A + \begin{pmatrix} I \\ 0 \end{pmatrix} E - \begin{pmatrix} X \\ 0 \end{pmatrix} \right\|_F^2, \\ T^{k+1} = T^k - \mu \left( \begin{pmatrix} V \\ I \end{pmatrix} Z^{k+1} + \begin{pmatrix} 0 \\ -I \end{pmatrix} A^{k+1} + \begin{pmatrix} I \\ 0 \end{pmatrix} E^{k+1} - \begin{pmatrix} X \\ 0 \end{pmatrix} \right). \end{cases} \quad (10)$$

Because of $\begin{pmatrix} 0 \\ -I \end{pmatrix}^T \times \begin{pmatrix} I \\ 0 \end{pmatrix} = 0$, $A^{k+1}$ and $E^{k+1}$ are separable, Eq. (10) can be extended to

$$\begin{cases} Z^{k+1} = \arg\min_Z L_\mu(Z, A^k, E^k; T^k), \\ A^{k+1} = \arg\min_A L_\mu(Z^{k+1}, A, E^k; T^k), \\ E^{k+1} = \arg\min_E L_\mu(Z^{k+1}, A^{k+1}, E; T^k), \\ T^{k+1} = T^k - \mu \left( \begin{pmatrix} 0 \\ -I \end{pmatrix} A^{k+1} + \begin{pmatrix} V \\ I \end{pmatrix} Z^{k+1} + \begin{pmatrix} I \\ 0 \end{pmatrix} E^{k+1} - \begin{pmatrix} X \\ 0 \end{pmatrix} \right). \end{cases} \quad (11)$$

For $Z^{k+1}$,

$$Z^{k+1} = \arg\min_Z \|Z\|_* + \frac{\mu}{2}\left\| \begin{pmatrix} V \\ I \end{pmatrix} Z + \begin{pmatrix} 0 \\ -I \end{pmatrix} A^k + \begin{pmatrix} I \\ 0 \end{pmatrix} E^k - \begin{pmatrix} X \\ 0 \end{pmatrix} - \frac{T^k}{\mu} \right\|_F^2$$

$$= \arg\min_Z \|Z\|_* + \frac{\mu}{2}\left\| \begin{pmatrix} VZ + E^k - X - \frac{T_1^k}{\mu_1} \\ Z - A^k - \frac{T_2^k}{\mu_2} \end{pmatrix} \right\|_F^2$$

$$= \arg\min_Z \|Z\|_* + \frac{\mu_1}{2}\left\| VZ + E^k - X - \frac{T_1^k}{\mu_1} \right\|_F^2 + \frac{\mu_2}{2}\left\| Z - A^k - \frac{T_2^k}{\mu_2} \right\|_F^2,$$

Let $F(Z) = \left\| VZ + E^k - X - \frac{T_1^k}{\mu_1} \right\|_F^2 + \left\| Z - A^k - \frac{T_2^k}{\mu_2} \right\|_F^2$. Where $T = \begin{pmatrix} T_1 \\ T_2 \end{pmatrix}$, $\mu = (\mu_1, \mu_2)$, $\mu_1, \mu_2 > 0$ are penalty parameters. $\nabla_Z F$ is the partial differentiation of function $F(Z)$ with respect to $Z$.

$$\nabla_Z F(Z^k) = \mu_1 V^T (VZ^k + E^k - X - \frac{T_1^k}{\mu_1}) + \mu_2 Z^k - \mu_2 A^k - T_2^k = B^k,$$

$$Z^{k+1} = \arg\min_Z \|Z\|_* + \langle B^k, Z - Z^k \rangle + \frac{\eta_V}{2} \|Z - Z^k\|_F^2$$

$$= \arg\min_Z \|Z\|_* + \frac{\eta_V}{2} \left\| Z - (Z^k - \frac{B^k}{\eta_V}) \right\|_F^2 \qquad (14)$$

$$= D_{\eta_V^{-1}}(Z^k - \frac{B^k}{\eta_V}), \quad Z^{k+1} \in P_\mho(Z).$$

where $\eta_V \geq \|\mu_1 V^T V + \mu_2 I\|_2^2$, $D$ is the singular value threshold operator [30]. In order to satisfy the nonnegative constraints $Z_{ii} = 0$, projection operators can be constructed as follows,

$$(P_\mho(Z))_{qi} = \begin{cases} Z_{qi}, & \text{if } q \neq i, \\ 0, & \text{if } q = i, \end{cases} \quad Z_{qi} \in Z,$$

where $(P_\mho(Z))_{qi}$ is each component of the set $P_\mho(Z)$. Then the restriction $Z_{ii} = 0$ converts $Z \in P_\mho(Z)$.

For $A^{k+1}$ and $E^{k+1}$,

$$A^{k+1} = \arg\min_{A} \lambda_1 \|A\|_1 + (T_2^k)^T A + \frac{\mu_2}{2} \|Z^{k+1} - A\|_F^2$$

$$= \arg\min_{A} \lambda_1 \|A\|_1 + \frac{\mu_2}{2} \left\| Z^{k+1} - A - \frac{T_2^k}{\mu_2} \right\|_F^2 \quad (15)$$

$$= S_{\frac{\lambda_2}{\mu_2}} \left( Z^{k+1} - \frac{T_2^k}{\mu_2} \right).$$

$$E^{k+1} = \arg\min_{E} \lambda_2 \|E\|_1 + \frac{\mu_1}{2} \left\| E + VZ^{k+1} - X - \frac{T_1^k}{\mu_1} \right\|_F^2$$

$$= S_{\frac{\lambda_2}{\mu_1}} (X - VZ^{k+1} + \frac{T_1^k}{\mu_1}). \quad (16)$$

In Eq. (15) and Eq. (16), $S$ is the soft thresholding operator [31].

In order to satisfy the nonnegative constraints $A_{ii} = 0$, projection operators can be constructed as follows,

$$(P_\mho(A))_{qi} = \begin{cases} A_{qi}, & \text{if } q \neq i, \\ 0, & \text{if } q = i, \end{cases} \quad A_{qi} \in A,$$

where $(P_\mho(A))_{qi}$ is each component of the set $P_\mho(A)$. Then the restriction $A_{ii} = 0$ converts $A \in P_\mho(A)$.

**Appendix B**

The convergence theorem and the corresponding lemmas are given below. In order to prove the convergence theorem, similar to [19], three lemmas are firstly given. The symbols used in lemmas and theorem are as follows: $W^k = (A^k, Z^k, E^k, T^k)$, $U^k = (A^k, Z^k, E^k)$, $V^k = (A^k, E^k, T^k)$, and $V = (A, E, T)$.

Suppose $A \in \chi_1$, $Z \in \chi_2$, $E \in \chi_3$, $T \in \chi_4$, then $\Omega = \chi_1 \times \chi_2 \times \chi_3 \times \chi_4$, $\vartheta = \chi_1 \times \chi_3 \times \chi_4$, and

$$\vartheta^* := \{V^* = (A^*, E^*, T^*) | W^* = (A^*, Z^*, E^*, T^*) \in \Omega^*\},$$

where $(A^*, Z^*, E^*, T^*)$ is a saddle point of the Lagrangian function of Eq. (8). Similar to [19], it is useful to characterize the model Eq. (8) by a variational inequality. More specifically, finding a saddle point of the Lagrange function of Eq. (8) is equivalent to solve the variational inequality problem: finding a $W^* \in \Omega^*$ such that

$$\text{VI}(\Omega, F, \theta): \theta(U) - \theta(U^*) + (W - W^*)^T F(W^*) \geq 0, \ \forall W \in \Omega, \quad (17a)$$

where

$$U = (A, Z, E)^T, \ W = (A, Z, E, T)^T, \ \theta(U) = \|Z\|_* + \lambda_1 \|A\|_1 + \lambda_2 \|E\|_1 \quad (17b)$$

and

$$F(W) = \begin{pmatrix} -\begin{pmatrix} 0 \\ -I \end{pmatrix}^T \\ -\begin{pmatrix} V \\ I \end{pmatrix}^T \\ -\begin{pmatrix} I \\ 0 \end{pmatrix}^T \\ \begin{pmatrix} V \\ I \end{pmatrix} Z + \begin{pmatrix} 0 \\ -I \end{pmatrix} A + \begin{pmatrix} I \\ 0 \end{pmatrix} E - \begin{pmatrix} X \\ 0 \end{pmatrix} \end{pmatrix}. \quad (17c)$$

Obviously, the mapping $F(.)$ defined in Eq. (17c) is monotone because it is affine with a skew-symmetric matrix.

**Lemma 3.1.** Let $W^{k+1}$ be generated by Eq. (11) from given $V^k$. Then we have

$$W^{k+1} \in \Omega, \ \theta(U) - \theta(U^{k+1}) + (W - W^{k+1})^T \{F(W^{k+1}) + Q(V^k - V^{k+1})\} \geq 0,$$

$$\forall W \in \Omega, \quad (18)$$

where

$$Q = \begin{pmatrix} -\mu \begin{pmatrix} 0 \\ -I \end{pmatrix}^T \begin{pmatrix} 0 \\ -I \end{pmatrix} & 0 & \begin{pmatrix} 0 \\ -I \end{pmatrix}^T \\ 0 & \mu \begin{pmatrix} V \\ I \end{pmatrix}^T \begin{pmatrix} I \\ 0 \end{pmatrix} & 0 \\ \begin{pmatrix} 0 \\ -I \end{pmatrix}^T & 0 & -\frac{1}{\mu} I \end{pmatrix}.$$

**Lemma 3.2.** Let $W^{k+1}$ be generated by Eq. (11) from given $V^k$. If $\begin{pmatrix} 0 \\ -I \end{pmatrix}^T \cdot \begin{pmatrix} I \\ 0 \end{pmatrix} = 0$,

then we have

$$W^{k+1} \in \Omega, \ \theta(U) - \theta(U^{k+1}) + (W - W^{k+1})^T \{F(W^{k+1}) + \mu P \begin{pmatrix} I \\ 0 \end{pmatrix}(E^k - E^{k+1})\}$$

$$\geq (V - V^{k+1})^T H(V^k - V^{k+1}), \ \forall W \in \Omega, \quad (19)$$

where

$$P = \left( \begin{pmatrix} 0 \\ -I \end{pmatrix}^T, \begin{pmatrix} V \\ I \end{pmatrix}^T, \begin{pmatrix} I \\ 0 \end{pmatrix}^T, 0 \right)^T, \text{ and } H = \begin{pmatrix} \mu \begin{pmatrix} 0 \\ -I \end{pmatrix}^T \begin{pmatrix} 0 \\ -I \end{pmatrix} & 0 & -\begin{pmatrix} 0 \\ -I \end{pmatrix}^T \\ 0 & \mu \begin{pmatrix} I \\ 0 \end{pmatrix}^T \begin{pmatrix} I \\ 0 \end{pmatrix} & 0 \\ -\begin{pmatrix} 0 \\ -I \end{pmatrix}^T & 0 & \frac{1}{\mu} I \end{pmatrix}.$$

In the next lemma, an important inequality is established based on the assertion in Lemma 3.2, which will play a vital role in convergence analysis.

**Lemma 3.3.** Let $W^{k+1}$ be generated by Eq. (11) from given $V^k$. If $\begin{pmatrix} 0 \\ -I \end{pmatrix}^T \cdot \begin{pmatrix} I \\ 0 \end{pmatrix} = 0$,

for $\forall W \in \Omega$, we have $\tilde{W}^k \in \Omega$ and

$$\theta(U) - \theta(\tilde{U}^k) + (W - \tilde{W}^k)^T F(\tilde{W}^k) \geq \frac{1}{2}(\|V - V^{k+1}\|_H^2 - \|V - V^k\|_H^2) + \frac{1}{2}\|V^k - V^{k+1}\|_H^2, \quad (20)$$

where

$$\tilde{W}^k = \begin{pmatrix} \tilde{A}^k \\ \tilde{Z}^k \\ \tilde{E}^k \\ \tilde{T}^k \end{pmatrix} = \begin{pmatrix} A^{k+1} \\ Z^{k+1} \\ E^{k+1} \\ T^{k+1} - \mu \begin{pmatrix} I \\ 0 \end{pmatrix}(E^k - E^{k+1}) \end{pmatrix} \quad \text{and} \quad \tilde{U}^k = \begin{pmatrix} \tilde{A}^k \\ \tilde{Z}^k \\ \tilde{E}^k \end{pmatrix}.$$

Now, we are ready to prove the convergence of Eq. (11) under condition $\begin{pmatrix} 0 \\ -I \end{pmatrix}^T \times \begin{pmatrix} I \\ 0 \end{pmatrix} = 0$.

**Theorem 3.1 (Convergence Theorem)** $\begin{pmatrix} 0 \\ -I \end{pmatrix}^T \times \begin{pmatrix} I \\ 0 \end{pmatrix} = 0$ for the model Eq. (8), let $\{A^k, Z^k, E^k, T^k\}$ be the sequence generated by the direct extension of ADMM Eq. (11). Then, we have:

(i) The sequence $V^k$ is contractive with respective to the solution of $\text{VI}(\Omega, F, \theta)$, i.e.,

$$\left\| V^{k+1} - V^* \right\|_H^2 \leq \left\| V^k - V^* \right\|_H^2 - \left\| V^k - V^{k+1} \right\|_H^2. \tag{21}$$

(ii) If the matrices $\left[ \begin{pmatrix} 0 \\ -I \end{pmatrix}, \begin{pmatrix} V \\ I \end{pmatrix} \right]$ and $\begin{pmatrix} I \\ 0 \end{pmatrix}$ are assumed to be full column rank, then the sequence $\{W^k\}$ converges to a KKT point of the Eq. (8).

Now, we pay attention to the case where $\begin{pmatrix} 0 \\ -I \end{pmatrix}^T \times \begin{pmatrix} I \\ 0 \end{pmatrix} = 0$ and show that it can also ensure the convergence of Eq. (10). And the proof of the Lemmas 3.1, 3.2, 3.3, please refer to [19].

To prepare for the proof, we need to make something clear. First, note that the update order of Eq. (10) at each iteration is $Z \to (A, E) \to T$ and then it repeats cyclically. Equivalently, we can update the variables via the order $Z \to A \to E \to T$ and thus have the following iterative form:

$$\begin{cases} Z^{k+1} = \arg\min_{Z} L_{\mu}(A^k, Z, E^k; T^k), & \text{(11a)} \\ A^{k+1} = \arg\min_{A} L_{\mu}(A, Z^{k+1}, E^k; T^k), & \text{(11b)} \\ E^{k+1} = \arg\min_{E} L_{\mu}(A^{k+1}, Z^{k+1}, E; T^k), & \text{(11c)} \\ T^{k+1} = T^k - \mu\left(\begin{pmatrix} 0 \\ -I \end{pmatrix} A^{k+1} + \begin{pmatrix} V \\ I \end{pmatrix} Z^{k+1} + \begin{pmatrix} I \\ 0 \end{pmatrix} E^{k+1} - \begin{pmatrix} X \\ 0 \end{pmatrix}\right). & \text{(11d)} \end{cases}$$

Similar to [19], prove the process is as follows. The following is a proof of Theorem 3.1

**Proof.** (i) The first assertion is straightforward based on lemma 3.3. Setting $W = W^*$ in lemma 3.3, we get

$$\frac{1}{2}(\|V^k - V^*\|_H^2 - \|V^{k+1} - V^*\|_H^2) - \frac{1}{2}\|V^k - V^{k+1}\|_H^2 \geq \theta(\tilde{U}^k) - \theta(U^*) + (\tilde{W}^k - W^*)^T F(\tilde{W}^k).$$

Obviously, the mapping $F(.)$ defined in Eq. (17c) is monotone because it is affine with a skew-symmetric matrix. From the monotonicity of $F$ and Eq. (17), it follows that

$$\theta(\tilde{U}^k) - \theta(U^*) + (\tilde{W}^k - W^*)^T F(\tilde{W}^k) \geq \theta(\tilde{U}^k) - \theta(U^*) + (\tilde{W}^k - W^*)^T F(W^*) \geq 0,$$

and thus Eq. (21) is proved. Clearly, Eq. (21) indicates that the sequence $\{V^k\}$ is contractive with respect to the solution set of $VI(\Omega, F, \theta)$, see e.g. [32].

(ii) To prove (ii), by the inequality Eq. (21) and (see the definitions of $V$ and $H$ in lemma 3.2)

$$\|V^k - V^{k+1}\|_H^2 = \mu\left\|\begin{pmatrix} 0 \\ -I \end{pmatrix}(A^k - A^{k+1}) - \frac{1}{\mu}(T^k - T^{k+1})\right\|^2 + \mu\left\|\begin{pmatrix} I \\ 0 \end{pmatrix}(E^k - E^{k+1})\right\|^2,$$

it follows that the sequences $\left\{\begin{pmatrix} 0 \\ -I \end{pmatrix} A^k - \frac{1}{\mu}T^k\right\}$ and $\left\{\begin{pmatrix} I \\ 0 \end{pmatrix} E^k\right\}$ are both bounded.

Since $\begin{pmatrix} I \\ 0 \end{pmatrix}$ has full column rank, we deduce that $\{E^k\}$ is bounded. Note that

$$\begin{pmatrix} 0 \\ -I \end{pmatrix} A^k + \begin{pmatrix} V \\ I \end{pmatrix} Z^k = \begin{pmatrix} 0 \\ -I \end{pmatrix} A^k - \frac{1}{\mu}T^k - \left(\begin{pmatrix} 0 \\ -I \end{pmatrix} A^{k-1} - \frac{1}{\mu}T^{k-1}\right) - \begin{pmatrix} I \\ 0 \end{pmatrix} E^k + \begin{pmatrix} X \\ 0 \end{pmatrix}. \quad (22)$$

Hence $\left\{\begin{pmatrix}0\\-I\end{pmatrix}A^k+\begin{pmatrix}V\\I\end{pmatrix}Z^k\right\}$ is bounded. Together with the assumption that $\left[\begin{pmatrix}0\\-I\end{pmatrix},\begin{pmatrix}V\\I\end{pmatrix}\right]$ has full column rank, we conclude that the sequences $\{A^k\}$, $\{Z^k\}$ and $\{T^k\}$ are all bounded. Therefore, there exists a subsequence $\{A^{n_k+1},Z^{n_k+1},E^{n_k+1},T^{n_k+1}\}$ that converges to a limit point, say $\{A^\infty,Z^\infty,E^\infty,T^\infty\}$. Moreover, from Eq. (21), we see immediately that

$$\sum_{k=1}^{\infty}\left\|V^k-V^{k+1}\right\|_H^2<+\infty,$$

which shows

$$\lim_{k\to\infty}H(V^k-V^{k+1})=0,$$

Similar to lemma 3.1, we defined

$$Q=\begin{pmatrix}-\mu\begin{pmatrix}0\\-I\end{pmatrix}^T\begin{pmatrix}0\\-I\end{pmatrix} & 0 & \begin{pmatrix}0\\-I\end{pmatrix}^T \\ 0 & \mu\begin{pmatrix}V\\I\end{pmatrix}^T\begin{pmatrix}I\\0\end{pmatrix} & 0 \\ \begin{pmatrix}0\\-I\end{pmatrix}^T & 0 & -\dfrac{1}{\mu}I\end{pmatrix},$$

and thus

$$\lim_{k\to\infty}Q(V^k-V^{k+1})=0. \tag{23}$$

Then, by taking the limits on the both sides of lemma 3.1, using Eq. (23), one can immediately write

$$W^\infty\in\Omega,\ \theta(U)-\theta(U^\infty)+(W-W^\infty)^T F(W^\infty)\geq 0,\ \forall W\in\Omega,$$

which means $W^\infty=(A^\infty,Z^\infty,E^\infty,T^\infty)$ is a KKT point of Eq. (10). Hence, the inequality Eq. (21) is also valid if $(A^*,Z^*,E^*,T^*)$ is replaced by $(A^\infty,Z^\infty,E^\infty,T^\infty)$. Then it holds that

$$\left\|V^{k+1}-V^\infty\right\|_H^2\leq\left\|V^k-V^\infty\right\|_H^2,$$

which implies that

$$\lim_{k \to \infty} \begin{pmatrix} 0 \\ -I \end{pmatrix}(A^k - A^\infty) - \frac{1}{\mu}(T^k - T^\infty) = 0, \quad \lim_{k \to \infty} \begin{pmatrix} I \\ 0 \end{pmatrix}(E^k - E^\infty) = 0. \tag{24}$$

By taking limits to Eq. (22), using Eq. (24) and the assumptions, we know

$$\lim_{k \to \infty} A^k = A^\infty, \quad \lim_{k \to \infty} Z^k = Z^\infty, \quad \lim_{k \to \infty} E^k = E^\infty, \quad \lim_{k \to \infty} T^k = T^\infty.$$

which completes the proof of this theorem.

The theorem illustrates if the iterative solution of model exists, the solution satisfies the constraint conditions and converges to the numerical solution. Futhermore, the convergence bound theorem is given below, where the theorem shows the convergence speed.

In order to give the theorem, similar to [20], we need to partition 3 blocks variables into 2 super blocks first. Assume that we give a partition of 3 blocks, denoted as $\{B_1, B_2\}$, we accordingly partition $C$ into $C_{B_1} = [C_i, i \in B_1]$ and $C_{B_2} = [C_i, i \in B_2]$. Then Eq. (8) is equivalent to

$$\min_{x_{B_1}, x_{B_2}} f(x),$$
$$s.t. \ C_{B_1} x_{B_1} + C_{B_2} x_{B_2} = d.$$

where $x_{B_1} = \{Z | Z_{ii} = 0\}$, $x_{B_2} = \{A, E | A_{ii} = 0\}$.

**Theorem 3.2 (Convergence Bound Theorem)**

In Algorithm 1, assume that $f^k \in S_{\{L_i, P_i\}_{i=1}^n}(f, x^k)$ [20] with $P_i \succeq L_i \succeq 0$, $\frac{1}{2}\|C_{B_1} x_{B_1}\|^2$ is $\{L_i'\}_{i \in B_1}$ -smooth [20], $\frac{1}{2}\|C_{B_2} x_{B_2}\|^2$ is $\{L_i'\}_{i \in B_2}$ -smooth [20],

$G_i \succeq L_i' - C_i^T C_i, \ i \in B_1$ in

$\frac{1}{2}\left\|C_i x_i + \sum_{j \in B_1, j \neq i} C_j x_j^k + C_{B_2} x_{B_2}^k - d + \frac{T^k}{\mu^{(k)}}\right\|^2 + \frac{1}{2}\|x_i - x_i^k\|_{G_i}^2 + e_i^k, i \in B_1$ with $e_i^k$'s satisfying

$\sum_{i \in B_1} e_i^k = 0$ , $G_i \succ L_i' - C_i^T C_i, \ i \in B_2$ in

$\frac{1}{2}\left\|C_i x_i + \sum_{j \in B_2, j \neq i} C_j x_j^k + C_{B_1} x_{B_1}^{k+1} - d + \frac{T^k}{\mu^{(k)}}\right\|^2 + \frac{1}{2}\|x_i - x_i^k\|_{G_i}^2 + e_i^k, i \in B_2$ with $e_i^k$'s

satisfying $\sum_{i \in B_2} e_i^k = -\frac{1}{2}\left\|C_{B_1}x_{B_1}^{k+1} + C_{B_2}x_{B_2}^{k} - d + \frac{T^k}{\mu^{(k)}}\right\|^2$. For any $K > 0$, let

$\bar{x}^K = \sum_{k=0}^{K} \gamma^{(k)} x^{k+1}$, with $\gamma^k = (\mu^{(k)})^{-1} \bigg/ \sum_{k=0}^{K} (\mu^{(k)})^{-1}$. Then

$$f(\bar{x}^K) - f(x^*) + \langle C^T T^*, \bar{x}^K - x^* \rangle + \frac{\mu^{(0)}\alpha}{2}\|C\bar{x}^K - d\|^2$$

$$\leq \frac{\sum_{j=1}^{2}\|x_{B_j}^* - x_{B_j}^0\|_{H_j^0}^2 + \|T^* - T^0\|_{H_3^0}^2}{2\sum_{k=0}^{K}(\mu^{(k)})^{-1}}$$

where $\bar{x}^K$ is a weighted sum of $x^k$'s, and $\alpha$, $H_1^0$, $H_2^0$, $H_3^0$ are as follows:

$$\alpha = \min\left\{\frac{1}{2}, \frac{\delta_{\min}^2(Diag\{C_i^T C_i + G_i, i \in B_2\} - C_{B_2}^T C_{B_2})}{2\|C_{B_2}\|_2^2}\right\},$$

$$H_1^0 = Diag\left\{\frac{1}{\mu^{(0)}}L_i + C_i^T C_i + G_i, i \in B_1\right\} - C_{B_1}^T C_{B_1},$$

$$H_2^0 = Diag\left\{\frac{1}{\mu^{(0)}}L_i + C_i^T C_i + G_i, i \in B_2\right\}, \text{ and } H_3^0 = (1/\mu^{(0)})^2 I.$$

The stability analysis of the LR-S-PFSRC model and corresponding classification stability theorem is given below.

**Theorem 3.3 (Classification Stability Theorem).** Suppose $x_{i_2} = x_{i_1} + \Delta(x_{i_1})$ or corresponding pseudo-full-space has disturbance, i.e., $V_{i_2} = V_{i_1} + \Delta(V_{i_1})$ and corresponding LR-S-PFSR $x_{i_1} = V_{i_1}Z_{i_1}$, $x_{i_2} = V_{i_2}Z_{i_2}$, if

$$\varepsilon = \max\left\{\frac{\|\Delta(x_{i_1})\|_2}{\|x_{i_1}\|_2}, \frac{\|\Delta(V_{i_1})\|_2}{\|V_{i_1}\|_2}\right\} \leq \frac{\varphi_n(V_{i_1})}{\varphi_1(V_{i_1})},$$

and $\sin(\theta) = \rho_{LS}/\|x_{i_1}\|_2 \neq 1$, where $\rho_{LS} = \|V_{i_1}Z_{LS} - x_{i_1}\|_2$, $Z_{LS} = \arg\min\|x_{i_1} - V_{i_1}Z_{i_1}\|_2$, then

$$\frac{\|Z_{i_2} - Z_{i_1}\|_2}{\|Z_{i_1}\|_2} \leq \varepsilon\left\{\frac{2\kappa_2(V_{i_1})}{\cos(\theta)} + \tan(\theta)\kappa_2(V_{i_1})^2\right\} + o(\varepsilon^2). \tag{25}$$

The detail proof of the Theorem 3.3, please refer to [12]. The similarity between the stability analysis in this paper and that in [12] is that their representation spaces both are pseudo-full-space. However, the difference lies in the different constraints imposed on the coefficient matrices, and then their coefficient matrices are different.